\pdfoutput=1
\documentclass[conference]{IEEEtran}
\IEEEoverridecommandlockouts

\usepackage{amsmath}
\usepackage{amssymb}
\usepackage{bm}
\usepackage{enumitem}
\usepackage{graphics}

\usepackage{subcaption}
\usepackage{tikz}
\usetikzlibrary{positioning, calc, fit, shapes,arrows}

\usepackage{ifthen}
\usepackage{graphicx}
\usepackage{algorithm}
\usepackage{algpseudocode}

\usepackage{mathtools}

\DeclarePairedDelimiter\floor{\lfloor}{\rfloor}

\usepackage{array,tabularx}

\newenvironment{equation_parameters*}
  {\par\vspace{\abovedisplayskip}\noindent
   \tabularx{\columnwidth}{>{$}l<{$} @{}>{${}}c<{{}$}@{} >{\raggedright\arraybackslash}X}}
  {\endtabularx\par\vspace{\belowdisplayskip}}

\usepackage{multirow}

\def\BibTeX{{\rm B\kern-.05em{\sc i\kern-.025em b}\kern-.08em
    T\kern-.1667em\lower.7ex\hbox{E}\kern-.125emX}}
\begin{document}

\title{Assurance Monitoring of Cyber-Physical Systems with Machine Learning Components}

\author{\IEEEauthorblockN{Dimitrios Boursinos}
\IEEEauthorblockA{\textit{Institute for Software Integrated Systems} \\
\textit{Vanderbilt University}\\
Nashville TN, USA \\
dimitrios.boursinos@vanderbilt.edu}
\and
\IEEEauthorblockN{Xenofon Koutsoukos}
\IEEEauthorblockA{\textit{Institute for Software Integrated Systems} \\
\textit{Vanderbilt University}\\
Nashville TN, USA \\
xenofon.koutsoukos@vanderbilt.edu}
}

\maketitle

\begin{abstract}
\label{sec:abstract}
Machine learning components such as deep neural networks are used extensively in Cyber-Physical Systems (CPS). However, they may introduce new types of hazards that can have disastrous consequences and need to be addressed for engineering trustworthy systems. 
Although deep neural networks offer advanced capabilities, they must be complemented by engineering methods and practices that allow effective integration in CPS. 
In this paper, we investigate how to use the conformal prediction framework for assurance monitoring of CPS with machine learning components. In order to handle high-dimensional inputs in real-time,
we compute nonconformity scores using embedding representations of the learned models.
By leveraging conformal prediction, the approach provides well-calibrated confidence and can 
allow monitoring that ensures a bounded small error rate while limiting the number of inputs for which an accurate prediction cannot be made.
Empirical evaluation results using the German Traffic Sign Recognition Benchmark and a robot navigation dataset demonstrate that the error rates are well-calibrated while the number of 
alarms is small. The  method is computationally efficient,
and therefore, the approach is promising for assurance monitoring of CPS.
\end{abstract}
\section*{Keywords}
\label{sec:keywords}
Cyber-physical systems, deep neural networks, assurance monitoring,  conformal prediction, traffic sign recognition, robot navigation
\section{Introduction}
\label{sec:introduction}

Cyber-Physical systems (CPS) can benefit by incorporating machine learning components that can 
handle the uncertainty and variability of the real-world. Typical components such as deep neural networks (DNNs)
can be used for performing various tasks such as perception of the environment. 
In autonomous vehicles, for example,  perception problems deal with making sense of the surroundings like recognizing correctly traffic signs.
However, such DNNs introduce new types of hazards that can have disastrous consequences and need to be addressed for engineering trustworthy systems. 
Although DNNs offer advanced capabilities, they must be complemented by engineering methods and practices that allow effective integration in CPS. 

 A DNN is designed using learning techniques that require
specification of the task, performance measure for evaluating how well the task is performed, and experience 
which typically includes training and testing data. Using the DNN during system operation 
presents challenges that must be addressed using innovative engineering methods. 
Perception of the environment is a functionality that is difficult to specify, and typically, specifications
are based on examples.
DNNs exhibit some nonzero error rate and the true error rate is unknown and only an estimate from a design-time 
statistical process is known.
Further, DNNs encode information in a complex manner and it is hard to reason about the encoding.
Non-transparency is an obstacle to monitoring because it is more difficult to have confidence that the 
model is operating as intended.

The objective is to complement the prediction of DNNs
 with a computation of confidence. We consider DNNs used for classification in CPS.
In addition to the class prediction, we compute set and confidence predictors using 
 the conformal prediction framework~\cite{balasubramanian2014conformal}. 
 We focus on computationally efficient algorithms that can be used for real-time monitoring. 
 An efficient and robust approach must ensure a small and well-calibrated error rate while limiting 
 the number of alarms. This enables the design of monitors which can ensure a bounded small error 
rate while limiting the number of inputs for which an accurate prediction cannot be made.

The proposed approach is based on conformal prediction (CP)~\cite{Vovk:2005:ALR:1062391,balasubramanian2014conformal}.
CP aims at associating reliable measures of confidence with set predictions for problems that include classification and
regression. An important feature of the CP framework is the calibration of the obtained confidence
values in an online setting which is very promising for real-time monitoring in CPS applications.
These methods can be applied for a variety of machine learning algorithms that include DNNs. 
The main idea is to test if a new input example
conforms to the training data set by utilizing a \textit{nonconformity measure} 
which assigns a numerical score indicating how different the input example 
is from the training data set. 
The next step is to define a $p$-value as the fraction of
observations that have nonconformity scores greater than or equal to the nonconformity scores of the training examples which is then used for estimating the confidence of the prediction for the test input. 
In order to use the approach online, inductive conformal prediction (ICP) has been developed for computational efficiency~\cite{balasubramanian2014conformal}. 
In ICP, the training dataset is split into the proper training dataset that is used for learning and 
a calibration dataset that is used to compute the predictions for given confidence levels. Existing methods rely on nonconformity measures computed using techniques such as $k$-Nearest Neighbors and Kernel Density Estimation 
and do not scale for high-dimensional inputs in CPS.

In this paper, we investigate the ICP framework for assurance monitoring of CPS with machine learning components.
The approach leverages ICP for providing predictions with well-calibrated confidence.
The main contribution is that in order to handle high-dimensional inputs in real-time,
we compute the nonconformity scores using the embedding representations of the learned DNN models.
We combine the confidence predictions with a monitor which ensures a bounded small error 
rate while limiting the number of inputs for which an accurate prediction cannot be made.

A second contribution of the paper is that it presents an empirical evaluation of the approach using two datasets for classification problems in CPS. The first dataset is the German Traffic Sign Recognition Benchmark (GTSRB) dataset~\cite{GTSRB_cite}. For this dataset, we use MobileNet which is a popular network architecture that provides low-latency and low-power models to meet the resource constraints of a variety of use cases~\cite{howard2017mobilenets}.  The second dataset is the Scitos-G5 robot navigation dataset~\cite{Dua:2019} for which we used a fully connected feedforward network architecture.
We implement various nonconformity functions and we investigate if they can be computed efficiently in real-time.  The significance level threshold is selected either to a very small value driven by the CPS requirements or is computed to minimize the number of predictions with multiple classes.
The empirical results demonstrate that the error rates are well-calibrated 
and the number of alarms is small. Hence, we can design real-time monitors which can ensure a 
bounded small error rate while limiting the number of inputs for which an accurate 
prediction cannot be made.

Related work on confidence estimation for different kind of machine learning models follows in Section~\ref{sec:related_work}. In Section~\ref{sec:problem} we formally define the problem we worked on. In Section~\ref{sec:background} there is background on ICP that is used by our approach described in Section~\ref{sec:approach}. Finally, we evaluate the performance of our suggested approach on two different applications in Section~\ref{sec:evaluation}.

\section{Related Work}
\label{sec:related_work}
Confidence and uncertainty estimation in neural networks has received considerable attention especially in the context of classification tasks in computer vision~\cite{Guo:2017:CMN:3305381.3305518}. 
Neural networks for classification typically use a softmax layer. 
Correctly classified examples tend to have greater maximum 
softmax probabilities and several techniques have been proposed for estimating the error rate.
However, the softmax probabilities may be 
overconfident even for incorrect classes~\cite{Guo:2017:CMN:3305381.3305518}. 
This is the case because the softmax probabilities do not represent the actual probability distribution of the prediction. 


In CPS with machine learning components, computing well-calibrated confidence measures for the predictions is essential for providing system assurance but also making the behavior interpretable by humans. Modern DNNs keep increasing in size and are able to learn complicated training sets. Several methods have been proposed to calibrate the output probabilities of the predictor. A promising and very efficient method is \emph{temperature scaling}~\cite{Guo:2017:CMN:3305381.3305518}. The softmax activation function is used after temperature scaling to compute calibrated probability values. Although the method is very effective for calibration, it cannot be used for real-time monitoring in a straightforward manner.
Applying the approach for real-time assurance monitoring requires choosing an appropriate threshold which 
ensures a small error rate while limiting the  number of input examples for which a confident prediction cannot be made. Other methods for calibration include Platt scaling \cite{Platt99probabilisticoutputs} and isotonic regression \cite{Zadrozny:2002:TCS:775047.775151}. A machine learning model is expected to perform better in tasks that it has been trained on. Based on this assumption, the approach presented in~\cite{Gu:2019:TSM:3302509.3311038} analyzes the training space and tries to find areas where more training data are required. A decision tree algorithm is used to split the training space to areas based on their data density and a prediction probability measure is assigned to each region depending on the computed density.

Another framework developed to produce well-calibrated confidence values is Conformal Prediction (CP)~\cite{Vovk:2005:ALR:1062391,Shafer:2008:TCP:1390681.1390693,balasubramanian2014conformal}. The conformal prediction framework can be applied to produce calibrated confidence values with a variety of machine learning algorithms with slight modifications. Using CP together with methods that require long running times, such as DNNs, is computationally inefficient. In \cite{papadopoulos2007conformal} the authors suggest a modified version of the CP framework, they call Inductive Conformal Prediction (ICP), that has less computatonal overhead and they evaluate the results using DNNs as undelying model. Deep $k$-Nearest Neighbors (DkNN) is an approach based on ICP for classification problems that uses the activations from all the hidden layers of a neural network as features to the ICP~\cite{papernot2018deep}. This is based on the assumption that when a DNN make a wrong prediction there is a specific hidden layer that generated intermediate results that lead to the wrong prediction. So taking into account all the hidden layers, we have better interpretability of the predictions in each step. Another popular machine learning model is the Desicion Trees. In \cite{johansson2013conformal}, the authors present an empirical investigation of decision trees as conformal predictor and analyzed the algorithm's split criterion effect on ICP. Similarly, there are evaluations using ICP together with random forests \cite{Bhattacharyya2013}, \cite{10.1007/978-3-642-16239-8_8} as well as SVMs \cite{MAKILI20111213}. In all the above implementations of ICP the probability estimation of the prediction or \textit{credibility} is used to produce a different prediction than the underlying algorithm.

Confidence bounds can also be generated for regression problems. In this case instead of sets of multiple candidate labels we have intervals around a point prediction that include the correct prediction with a desired confidence. There are ICP implementations that work on regression problems with different underlying machine learning algorithms. In \cite{papadopoulos2011regression}, the authors use the $k$-Nearest Neighbours Regression ($k$-NNR) as a predictor and evaluate the effects of different nonconformity functions. Random forests can also be used in regression problems. In \cite{johansson2014regression}, there is comparison on the generated confidence bounds using $k$-NNR and DNNs \cite{papadopoulos2011reliable}. An alternative framework used to compute confidence bounds on regression problems is the Simultaneous Confidence Bands. In \cite{10.2307/2242228} they generate linear confidence bounds centered around the point prediction of a regression model. In this approach the model used for predictions has to be estimated by a sum of linear models. Models that satisfy this condition are the least squares polynomial models, kernel methods and smoothing splines. Functional Principal Components (FPC) analysis can be used for the decomposition of an arbitrary regression model to a combination of linear models \cite{goldsmith2013corrected}.

\section{Problem formulation}
\label{sec:problem}

A perception component in a CPS aims to observe and interpret the environment, in order to provide information for decision making.
For example, a DNN can be used for classifying traffic signs in autonomous vehicles. 
The problem is to complement the prediction of the DNN
 with a computation of the confidence. 
 An efficient and robust approach must ensure a small and well-calibrated error rate while limiting 
 the number of alarms to enable real-time monitoring. The approach must ensure a bounded small error 
rate while limiting the number of inputs for which an accurate prediction cannot be made. 

During the system operation, the inputs arrive one by one. After receiving each input,
the objective is to compute a valid measure of the confidence of the prediction.
The objective is twofold: (1) provide guarantees for the error rate of the prediction and 
(2) design a monitor which limits the number of input examples for which a confident prediction 
cannot be made. 
Such a monitor can be used for decision making, for example, by generating warnings and 
requiring human intervention.

The conformal prediction framework allows computing set and confidence predictors with well-calibrated confidence 
values~\cite{balasubramanian2014conformal}. The confidence is generated by comparing how similar a test is to the training data though different nonconformity functions. Our approach uses DNNs to generate a lower-dimensional embedding for each data point and estimates the similarity between different data points in the embedding space depending on the chosen nonconformity function. Depending on the chosen confidence bound the conformal prediction framework generates a set of possible predictions. If the computed set contains a single prediction, the confidence is a well-calibrated and valid indication of the expected 
error. If the computed set contains multiple predictions or no predictions, an alarm can be raised
to indicate the need for additional information.
In CPS, it is desirable to minimize the number of alarms while performing the required computations in real-time.
Evaluation of the method must be based on metrics that quantify the error rate, the number of alarms,
and the computational efficiency. For real-time operation, the time and memory requirements of the monitoring approach 
must be similar to the requirements of the DNNs used in the CPS architecture.

\section{Background}
\label{sec:background}

In this section, we give a brief overview of Inductive Conformal Prediction (ICP)~\cite{balasubramanian2014conformal} focusing on the definitions and notation necessary 
for presenting the monitoring algorithm.
Consider a training set \{$z_1,\dots,z_l$\} of examples, where each $z_i\in Z$ 
is a pair $(x_i,y_i)$ with $x_i$ the feature vector and $y_i$ the label of that example. 
Given an unlabeled input $x_{l+1}$, the task is to estimate a measure of confidence for 
different values $\tilde{y}$ for the label $y_{l+1}$ of this example. The underlying assumption 
for computing such measure of confidence is that all examples ($x_i,y_i$), $i=1,2,\dots$ are 
independent and identically distributed (IID) generated from the same but typically unknown 
probability distribution.

Essential in the application of the ICP is the definition of a \textit{nonconformity measure} which shows how different a labeled input is from the training examples. A nonconformity function 
assigns a numerical score to each example $z_i$ indicating how different the example $z_i$ is from the examples in  $\{z_1,\dots,z_{i-1},z_{i+1},\dots,z_n\}$.
The computation of the nonconformity is associated with an \textit{underlying algorithm} which maps an unlabeled example $x$ to the predicted label $\hat{y}$. 
There are many possible functions that can be used~\cite{balasubramanian2014conformal}.
A simple example is to count the number of the $k$-nearest neighbors to $z_{l+1}$ in $Z$ with label different than the candidate label $\tilde{y}$ 
($k$-nearest neighbors nonconformity measure).

Although lower nonconformity scores seem to correspond to higher confidence in the prediction, it is not possible to quantify the confidence based on absolute nonconformity scores. In order to compute a confidence measure, ICP uses a calibration dataset ($X^c,Y^c$). Using a nonconformity function, we can compute the nonconformity scores for all examples in the calibration data set 
\begin{equation}
A=\{\alpha(x,y):(x,y)\in(X^c,Y^c)\}.
\end{equation}
For a test example with feature vector  $ x $ and a candidate prediction $ j $, 
the nonconformity can be computed 
similarly to the calibration examples.
In order to compute useful predictions for test examples, 
ICP computes the fraction of nonconformity scores for the calibration data that are equal or larger than the nonconformity score of a test input. 
These are the empirical $p$-values for the test example defined as
\begin{equation}
\label{eq:p_values_equation}
p_j(x)=\frac{|\{\alpha\in A:\alpha\geq\alpha(x,j)\}|}{|A|}.
\end{equation}
Then, a set prediction $ \Gamma^\epsilon $ for the input $ x $ can be computed as the set of all labels $ j $ 
such that $ p_j(x) > \epsilon $. 

It is shown in~\cite{balasubramanian2014conformal} that predictors computed by ICP are valid, that is the probability of error will not exceed
$ \epsilon $ for any $\epsilon \in (0,1) $ for any choice of the nonconformity function. The problem is to compute
efficient predictors that output small prediction sets. In the case of real-time monitoring of CPS, computational efficiency is an additional requirement.

\section{Assurance Monitoring }
\label{sec:approach}

\subsection{Monitoring Algorithm}
\begin{figure}[ht]
\centering
\tikzstyle{block} = [draw, fill=blue!20, rectangle, 
    minimum height=3em, minimum width=3em]
\tikzstyle{input} = [coordinate]
\tikzstyle{output} = [draw=none, fill=none, circle, node distance=1cm]
\tikzstyle{pinstyle} = [pin edge={to-,thin,black}]
\tikzstyle{branch}=[fill,shape=circle,minimum size=3pt,inner sep=0pt]

\begin{tikzpicture}[auto, node distance=2cm,>=latex']
    \node [input, name=input] {};
    \node [block, right of=input,node distance=2cm] (dnn) {DNN};
    \node [block, below of=dnn,  node distance=1.8cm] (ICP) {ICP};
    \node [block, below of=ICP,  node distance=2.2cm] (assurance_monitor) {Assurance Monitor};
    \node [output, right of=dnn,node distance=2cm] (dnn_out) {$\hat{y}$};
    \node [output, below of=assurance_monitor,node distance=2cm] (assurance_out) {out};

    \draw [draw,->] (input) -- node {$x$} (dnn);
    \draw [->] (dnn) -- node [name=output] {}(dnn_out);
    \draw[->] (dnn) -- (ICP);
    \draw[->] (ICP) -- node [name=output] {$\Gamma^\epsilon$}(assurance_monitor);
    \draw[->] (assurance_monitor) -- (assurance_out);

\end{tikzpicture}
\caption{System Architecture}
\label{fig:problem_diagram}
\end{figure}
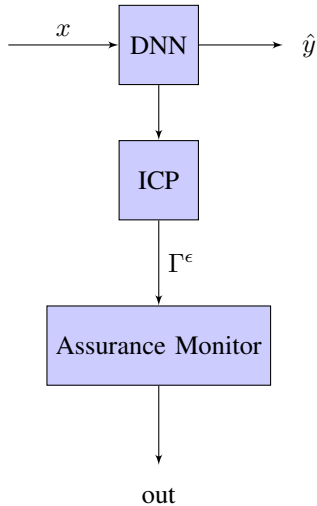

In CPS, we would like to design a monitoring algorithm which after receiving each input
computes a valid prediction that ensures a predefined error rate and limits the number of 
input examples for which a confident prediction cannot be made. Figure~\ref{fig:problem_diagram} illustrates the approach.
After receiving an input $ x $, the DNN is used not only to output a point prediction but also
to provide representations for efficiently computing the nonconformity scores for all possible
labels, which in turn, are used to compute a set prediction $ \Gamma^\epsilon $ at a given significance level
$ \epsilon $. The output of the monitor is defined as
\begin{equation*}
    out =
        \begin{cases}
            0,& \text{if } |\Gamma^\epsilon|=0\\
            1,& \text{if } |\Gamma^\epsilon|=1\\
            \text{reject},& \text{if } |\Gamma^\epsilon|>1
        \end{cases}
\end{equation*}
If the predicted set contains a single prediction, the monitor outputs $ out = 1 $ to indicate a confident prediction with well-calibrated error rate $ \epsilon $. If the predicted set contains multiple possible predictions, the monitor rejects the prediction and raises an alarm. Finally, If the predicted set is empty the monitor outputs $ out = 0 $ to indicate that no label is probable. We distinguish between multiple and no predictions, because they may lead to different action in the system. For example, no prediction may be the result of out-of-distribution inputs. The algorithm is shown in Algorithm \ref{alg:approach}.

\begin{algorithm}[ht]
	\caption{\textbf{-- Monitoring Algorithm}.}
	\label{alg:approach}
	\begin{algorithmic}[1] 
		\Require training data $(X, Y)$, calibration data $(X^c, Y^c)$
		\Require trained neural network $f$ with $l$ layers
		\Require Nonconformity function $\alpha$
		\Require test input $z$
		\Require significance level threshold $\epsilon$
		\State // Compute the nonconformity scores for the calibration data offline
		\State $A=\{\alpha(x,y) : (x, y) \in (X^c, Y^c)\}$ \Comment{Calibration}
		\State // Generate prediction sets for each test data online
		\For{each label $j\in 1..n$}
		\State Compute the nonconformity score $\alpha(z,j)$
		\State $p_j(z) = \frac{\left| \{ \alpha \in A : \alpha \geq \alpha(z, j) \} \right|}{|A|}$ \Comment{empirical $p$-value}
		\If{$p_j(z)\geq\epsilon$}
		\State Add $j$ to the prediction set $\Gamma^\epsilon$
		\EndIf
		\EndFor
		\If{$|\Gamma^\epsilon|=0$}
		\State \Return 0
		\ElsIf{$|\Gamma^\epsilon|=1$}
		\State \Return 1
		\Else
		\State\Return Reject
		\EndIf
	\end{algorithmic}
\end{algorithm}

\subsection{Low-dimensional Learned Representations}
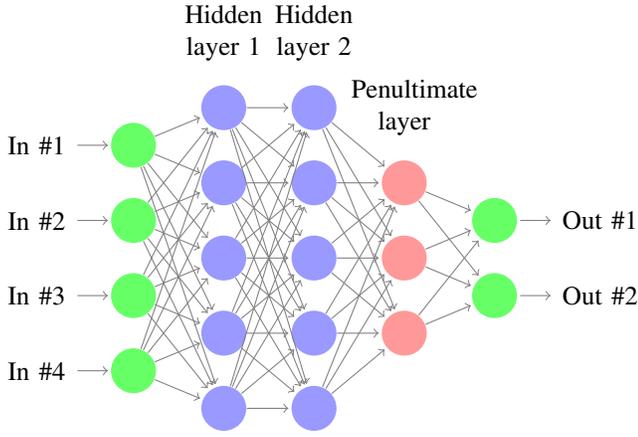
\begin{figure}[ht]
\def\layersep{1.2cm} 
\def\ha{4} 
\def\hb{5} 
\def\hc{5} 
\def\hd{3} 
\def\ho{2} 
\begin{tikzpicture}[shorten >=1pt,->,draw=black!50, node distance=\layersep]
    \tikzstyle{every pin edge}=[<-,shorten <=1pt]
    \tikzstyle{neuron}=[circle,fill=black!25,minimum size=17pt,inner sep=0pt]
    \tikzstyle{input neuron}=[neuron, fill=green!60];
    \tikzstyle{output neuron}=[neuron, fill=green!60];
    \tikzstyle{hidden neuron}=[neuron, fill=blue!40];
    \tikzstyle{embedding neuron}=[neuron, fill=red!40];
    \tikzstyle{annot} = [text width=4em, text centered]

    \foreach \name / \y in {1,...,\ha}
        \node[input neuron, pin=left:In \#\y] (I-\name) at (0,-\y) {};

    \foreach \name / \y in {1,...,\hb}
        \path[yshift=0.5cm]
            node[hidden neuron] (H1-\name) at (\layersep,-\y cm) {};
            
    \foreach \name / \y in {1,...,\hc}
        \path[yshift=0.5cm]
            node[hidden neuron] (H2-\name) at (2*\layersep,-\y cm) {};
            
    \foreach \name / \y in {1,...,\hd}
        \path[yshift=-0.5cm]
            node[embedding neuron] (E-\name) at (3*\layersep,-\y cm) {};
            
    \foreach \name / \y in {1,...,\ho}
        \path[yshift=-1.0cm]
            node[output neuron, pin={[pin edge={->}]right:Out \#\y}] (O-\name) at (4*\layersep,-\y) {};


    \foreach \source in {1,...,\ha}
        \foreach \dest in {1,...,\hb}
            \path (I-\source) edge (H1-\dest);
            
    \foreach \source in {1,...,\hb}
        \foreach \dest in {1,...,\hc}
            \path (H1-\source) edge (H2-\dest);
            
    \foreach \source in {1,...,\hc}
        \foreach \dest in {1,...,\hd}
            \path (H2-\source) edge (E-\dest);
            
    \foreach \source in {1,...,\hd}
        \foreach \dest in {1,...,\ho}
            \path (E-\source) edge (O-\dest);


    \node[annot,above of=H1-1, node distance=1cm] (hl) {Hidden layer 1};
    \node[annot,above of=H2-1, node distance=1cm] (hl) {Hidden layer 2};
    \node[annot,above of=E-1, node distance=1cm] (hl) {Penultimate layer};

\end{tikzpicture}
\caption{DNN architecture}
\label{fig:embedding_schematic}
\end{figure}

Typical nonconformity measures, such as the $k$-nearest neighbor ($k$-NN) nonconformity measure, are 
computed by considering the input space of the underlying algorithm. Perception components for CPS
use high-dimensional inputs such as images or LiDAR point clouds. For such cases, we investigate if
we can use nonconformity functions that are computed using low-dimensional representations learned by the DNN.
In particular, we use the activations of a fully connected penultimate layer to extract feature representations 
from the inputs (Fig.~\ref{fig:embedding_schematic}).

The embedding of inputs such as images reduces the dimensionality of the input data and allows the efficient computation of the nonconformity measure. In addition, we can use Euclidean distance in the corresponding vector space to compute informative nonconformity measures that lead to efficient predictions.
Ideally, the representation of a test input will be closer to representations of the same class and far from representations of different classes. We experiment with different number of neurons for the penultimate layer and we evaluate the effect on the performance and computational efficiency of the approach. A promising research direction for future work is to learn representations that lead to better set and confidence predictors. 

\subsection{Nonconformity Measure}
There are different nonconformity functions that can be used to evaluate how unusual a specific input is relative to the training set. 
We organize the nonconformity functions based on the features of the underlying model they use.

\subsubsection{Penultimate Layer}
A natural choice of the nonconformity function is how much the prediction of the underlying algorithm differs from the labels of the closest neighbors. We compute the \textit{$k$-Nearest Neighbors} ($k$-NN) nonconformity function in the space defined by the lower-dimensional penultimate layer which can reduce the required memory for the storage of the training data as well as the execution time. Let us denote by $f: X\rightarrow V$ the mapping from the input space $ X $ to the space $ V $ defined the penultimate layer enconding. After training is complete, we compute and store the encodings $v_i=f(x_i)$ for the training data $ x_i $. Given a test example $ x $ with encoding $ v = f(x) $, we compute the $k$-nearest neighbors in $ V $ and form a multi-set $\Omega$ with their labels. The $k$-NN nonconformity of the test example $ x $ with the candidate label $y$ is defined as:
\begin{equation*}
\label{eq:nonconformity_multiple_neighbors}
\alpha(x,y)=|i\in\Omega:i\neq y|.
\end{equation*}

Another nonconformity function is the \textit{one nearest neighbor} (1-NN) function which evaluates how far the closest training example of the same class is compared to the closest training example in any other class~\cite{Vovk:2005:ALR:1062391} and can be defined as 
\begin{equation*}
\label{eq:nonconformity_1neighbor}
\alpha(x,y)=\dfrac{\min_{i=1,\dots,n:y_i=y}d(v,v_i)}{\min_{i=1,\dots,n:y_i\neq y}d(v,v_i)}
\end{equation*}
where $ v = f(x) $, $v_i=f(x_i)$, and $d$ is a distance metric in the $V$ space.

These nonconformity functions require storing the training data set. Since we expect similar inputs of the same class to be close together and farther from inputs of other classes, we can also use the \emph{nearest centroid nonconformity function}~\cite{balasubramanian2014conformal}. For each class $y_i$ we compute its centroid $\mu_{y_i}=\frac{\sum_{j=1}^{n_i}v_j^i}{n_i}$, where $v_j^i$ is the representation of the  $ j^{th} $ training example from class $y_i$ and $n_i$ is the number of training examples in class $y_i$. The nonconformity function is defined as:
\begin{equation*}
\label{eq:nonconformity_nearest_centroid}
\alpha(x,y)=\dfrac{d(\mu_y,v)}{\min_{i=1,\dots,n:y_i\neq y}d(\mu_{y_i},v)}
\end{equation*}
where $ v = f(x) $ and we need to store only the centroid for each class.

\subsubsection{Softmax Layer}
A class of nonconformity functions can be computed using only the activations of the softmax layer~\cite{model_agnostic}. This class does not require storing information 
related to the training data, and thus, can be used for real-time monitoring. The softmax activation function $\sigma_{SM}$ normalizes the outputs of the last layer to empirical probabilities $\hat{p}_i=\hat{P}(y_i|x)$ that sum to 1~\cite{Guo:2017:CMN:3305381.3305518}. 

Three nonconformity functions, \textit{hinge}, \textit{margin} and \textit{brier score}, are suggested in~\cite{model_agnostic}.
Using the \textit{hinge} function, the nonconformity is computed using the probability estimate of the candidate class label, $y$
\begin{equation*}
\label{eq:nonconformity_hinge}
\alpha(x,y)=1-\hat{P}(y|x).
\end{equation*}
The \textit{Margin} function considers two class labels, the candidate
class label $y$ and the most likely incorrect class label $y_i$
\begin{equation*}
\label{eq:nonconformity_margin}
\alpha(x,y)=\max_{y_i\neq y}\hat{P}(y_i|x)-\hat{P}(y|x)
\end{equation*}
The \textit{Brier score nonconformity function} considers all the computed softmax probabilities $\hat{p}_i$ of a test input and it computes the nonconformity scores by comparing $\hat{p}_i$ with $P(y_i|x)$ assuming that for the candidate label $P(y|x)=1$ and for all the other labels $P(y_i|x)=0$
\begin{equation*}
\label{eq:nonconformity_brier_score}
\alpha(x,y)=\dfrac{1}{|Y|}\sum_{i=1}^{|Y|}({P}[y_i|x]-\hat{P}[y_i|x])^2
\end{equation*}
where $ Y $ is the set of all classes.

The nonconformity scores for the calibration data are computed using the ground truth labels. For test examples, the nonconformity scores are computed for every candidate class.  A candidate class is included in the set prediction $ \Gamma^\epsilon$ if the corresponding $p$-value is greater than the significance level $\epsilon$.

The empirical probabibilities computed using the softmax layer may not be well-calibrated.  \textit{Temperature scaling} is a simple method to calibrate neural networks~\cite{Guo:2017:CMN:3305381.3305518}. The probabilities can be computed as  $\hat{q}_i=\hat{Q}(y_i|x)=\sigma_{SM}(z_i/T)$, where $\bm{z}$ is the logits vector and $T$ is a variable called temperature. $T$ is computed by optimizing the negative log loss (NLL) on a validation set. The hinge, margin, and Brier score nonconformity functions can be combined with temperature scaling to compute the temparature scaled (TS) hinge, margin, and Brier score nonconformity functions respectively as 
$$\alpha(x,y)=1-\hat{Q}(y|x),$$
$$\alpha(x,y)=\max_{y_i\neq y}\hat{Q}(y_i|x)-\hat{Q}(y|x),$$ and 
$$\alpha(x,y)=\dfrac{1}{|Y|}\sum_{i=1}^{|Y|}({Q}[y_i|x]-\hat{Q}[y_i|x])^2.$$

\subsection{Significance Level Threshold}
In CPS, it is not only essential to have a well-calibrated confidence for a prediction but also to control the significance level that affects the risk of incorrect predictions. 
For a safety critical system, ideally the significance level $\epsilon$ could be selected to be $0$. However, in this case the set predictor will return all classes as possible. In CPS, the significance level $\epsilon$ can be selected based on the requirements of the application to ensure a desirable rate. In this case, we assume that set predictions with multiple classes, 
i.e. $|\Gamma^\epsilon| > 1$, lead to a rejection of the input and require human intervention. In this case, it is desirable to minimize the number of test inputs with multiple predictions.

Alternatively, we can select $\epsilon$ to the smallest value that aims to eliminate test inputs with multiple predictions. Given a validation set, we compute the number of predictions with multiple classes for different values of $\epsilon$ and we select the value that produced the minimum number.

\section{Evaluation}
\label{sec:evaluation}
The objective of the evaluation is to compare the validity and efficiency (size of set predictions) as well as the computational efficiency of the monitoring algorithm for different nonconformity functions as well as a baseline ICP approach that takes place in the input space.

\subsection{Experimental Setup}
For the experiments, we use two datasets. First, the German Traffic Sign Recognition Benchmark (GTSRB) dataset is a collection of traffic sign images to be classified in 43 classes (each class corresponds to a type of traffic sign)~\cite{GTSRB_cite}. It has 26640 labeled images of various sizes between 15x15 to 250x250 depending on the distance of the traffic sign to the vehicle. We convert all the images to 32x32 pixels. For this dataset, MobileNet, a popular Convolutional Neural Network (CNN) architecture that provides low-latency and low-power models, is used as the network architecture~\cite{howard2017mobilenets}. We use width multiplier $\alpha=1$ for the convolutional layers and a fully connected penultimate layer of size 128 is used to compute the encodings
for the $k$-NN, $1$-NN, and nearest centroid nonconformity functions. 

The second dataset is the SCITOS-G5 wall following robot navigation dataset~\cite{Dua:2019}. This dataset contains 5456 raw values of the measurements of 24 ultrasound sensors of a robot that are used to select actions ("Move-Forward", "Sharp-Right-Turn","Slight-Left-Turn", and "Slight-Right-Turn") so that the robot stays close to the wall.
Since the inputs in the SCITOS-G5 dataset come from 24 sensors, we treat them as vectors and use a fully connected neural network with one hidden layer. The number of hidden units, $h = 20$, is selected using a simple rule of thumb $h=\floor*{\frac{2a}{3}+C}$, where $a =24$ is the number of inputs and $C$ the number of classes~\cite{model_agnostic}. The penultimate layer that is used to compute the encodings is the single hidden layer.
For the baseline ICP application we use the $k$-NN, $1$-NN, and nearest centroid nonconformity functions applied input space instead of an embedding space to see if the embedding space improves the validity and efficiency in the CPS domain.

For each dataset, we use 10\% of the available data for testing. From the the rest 90\% of the data, 80\% is used for training and 20\% for calibration and/or validation. For the $k$-NN nonconformity function, we use $ k=15$.
All the experiments run in a desktop computer equipped with and Intel(R) Core(TM) i9-9900K CPU and 32 GB RAM and a Geforce RTX 2080 GPU with 8 GB memory.

\subsection{Assurance Monitoring}
\begin{figure}[htb]
\centering
\includegraphics[width=0.85\linewidth]{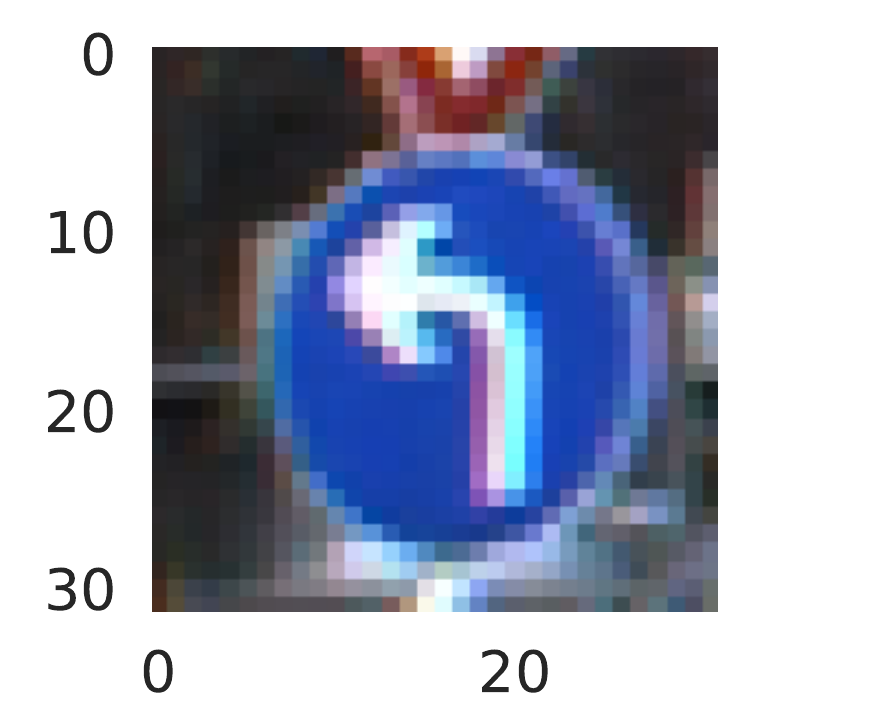}
\caption{Left turn sign}
\label{fig:sign_example}
\end{figure}

\begin{figure}[ht]
    \centering
    \begin{subfigure}[b]{4cm}
        \includegraphics[width=\textwidth,height=\textwidth]{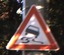}
        \caption{Attention slippery}
        \label{fig:slip_road}
    \end{subfigure}
    \hfill
    \begin{subfigure}[b]{4cm}
        \includegraphics[width=\textwidth,height=\textwidth]{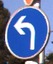}
        \caption{Turn left}
        \label{fig:turn_left}
    \end{subfigure}
    
    ~ 
    \begin{subfigure}[b]{4cm}
        \includegraphics[width=\textwidth,height=\textwidth]{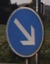}
        \caption{Turn right down}
        \label{fig:turn_right_down}
    \end{subfigure}
    \hfill
    \begin{subfigure}[b]{4cm}
        \includegraphics[width=\textwidth,height=\textwidth]{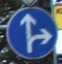}
        \caption{Turn straight right}
        \label{fig:turn_straight_right}
    \end{subfigure}
    \caption{The candidate classes for the example left turn sign input}
    \label{fig:candidate_labels}
\end{figure}

First, we illustrate the assurance monitoring algorithm with a test example from the GSTRB dataset. Figure~\ref{fig:sign_example} shows the image of a left turn sign. Using $k$-NN as the nonconformity function, Algorithm~\ref{alg:approach} can be used to generate sets of possible predicted labels. In the following, we vary the significance level $ \epsilon $ and we report the set predictions. 
When $\epsilon \in [0.001,0.003)$, the possible labels are 'attention\_slippery', 'turn\_left', 'turn\_straight\_right', 'turn\_right\_down'; when $\epsilon \in [0.003,0.018)$, the possible labels are 'turn\_left',  'turn\_right\_down'; and finally when $\epsilon\in$[0.018,0.1], the algorithm produces a single prediction 'turn\_left' which is obviously correct. The images of the signs in the above candidate classes can be seen in Figure~\ref{fig:candidate_labels}.

As expected, small values of the significance level increase the number of multiple predictions. The reason is that labels corresponding to images with some similarity are also considered possible. Larger significance levels generate single valued sets, however, the error rate is also higher. 

\subsection{Performance and Calibration}
The next goal is to evaluate the performance of the algorithm using testing datasets. First, the DNN model is trained and early stopping is used to reduce overfitting. The mobileNet used for the GTSRB dataset has training accuracy 98.1\% and test accuracy 96.5\%. The simple feedforward neural network used for the SCITOS-G5 navigation dataset has training accuracy 88.5\% and test accuracy 86.8\%.

\begin{figure}[ht]
\centering
\includegraphics[width=\linewidth]{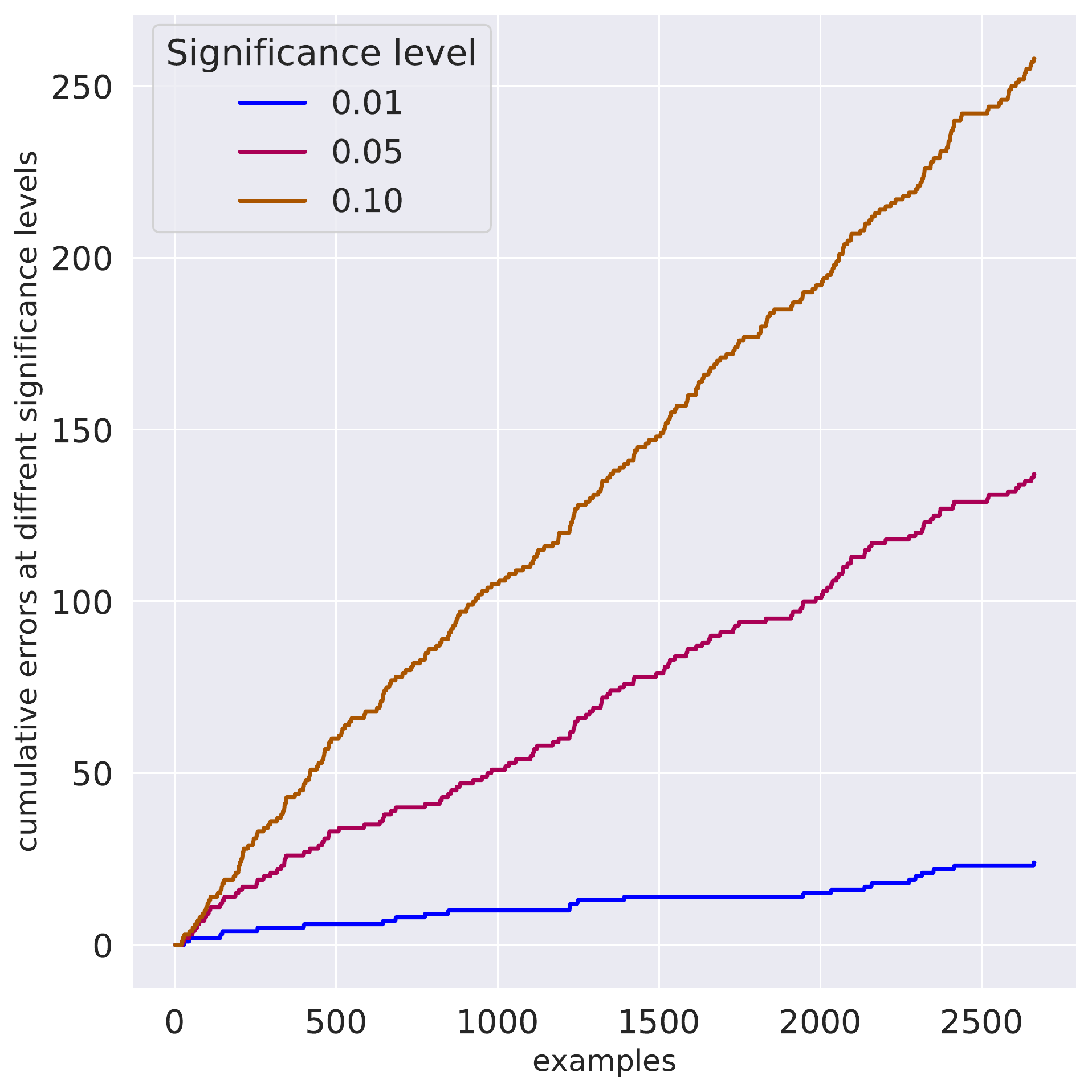}
\caption{Cumulative error curve}
\label{fig:errors_curve_plot}
\end{figure}

We would like to verify that the conformal prediction framework results is well-calibrated measures of confidence for the selected nonconformity functions and the error rate of the monitoring algorithm is bounded by the significance level. We compute the percentage of incorrect predictions and we plot the cumulative error for different values of $\epsilon$. In Figure \ref{fig:errors_curve_plot}, we plot the cumulative error for three different values of $\epsilon$ for the GTSRB dataset using the Nearest Centroid nonconformity function. The results show that the error rate is bounded by $\epsilon$. Similar behavior is observed using the other nonconformity functions.

\begin{figure}[ht]
\centering
\includegraphics[width=\linewidth]{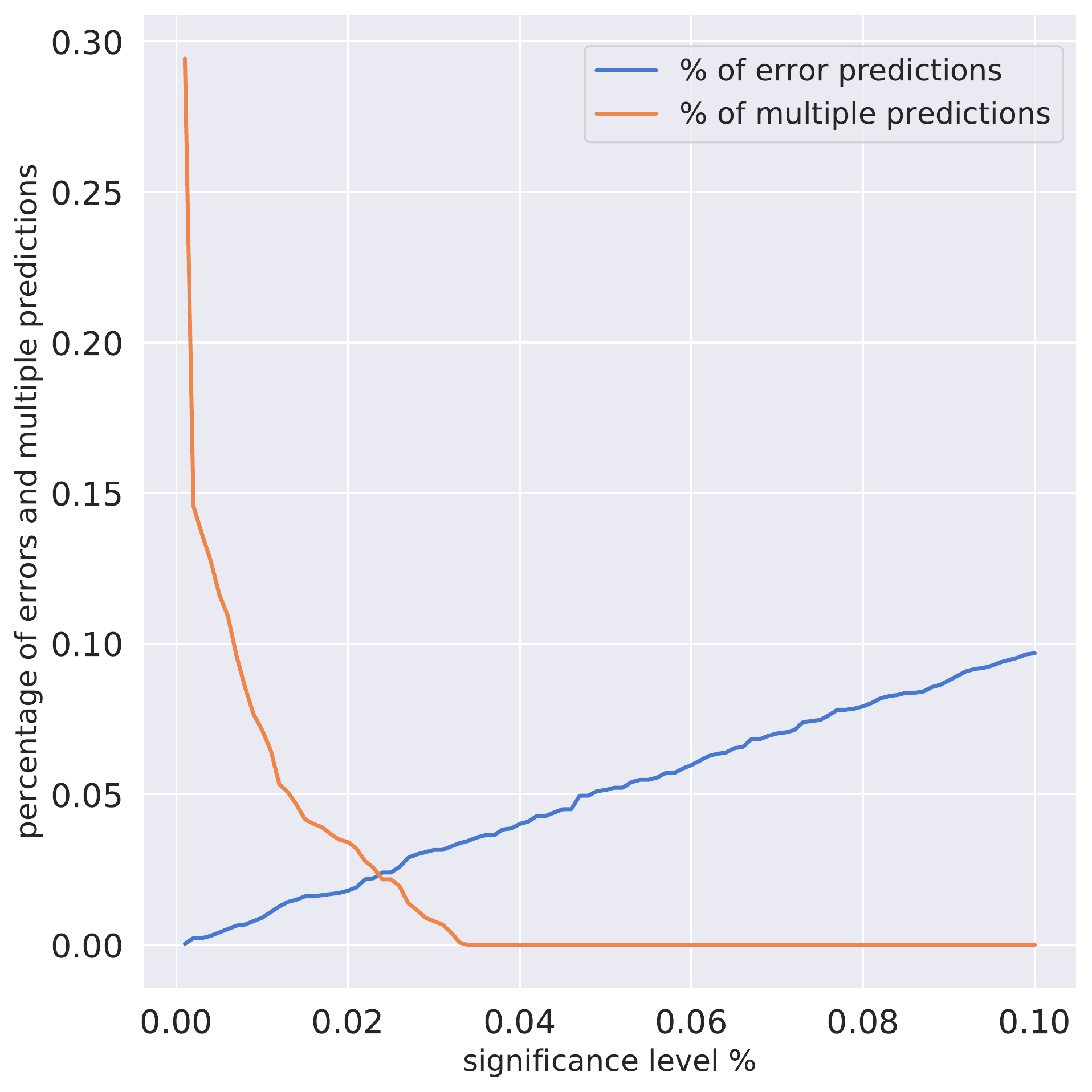}
\caption{Calibration and performance curve}
\label{fig:performance_curve_plot}
\end{figure}

In addition, we evaluate the performance and calibration of the obtained confidence values. 
Figure~\ref{fig:performance_curve_plot} shows the performance (\% of multiple predictions) and calibration (\% of error prediction) curve when $\epsilon\in [0.001, 0.1] $ for the GTSRB dataset using the Nearest Centroid nonconformity function. The number of multiple predictions decreases fast as $\epsilon$ increases. Further, the error rate is well-calibrated model and increases linearly with $\epsilon$. 

\subsection{Selecting the Significance Level}
For monitoring of CPS, one can either choose $\epsilon$ to be small enough given the system requirements or compute $ \epsilon $ to minimize the number of multiple predictions.
Since the number of multiple prediction decreases when $\epsilon$ increases, we can select $\epsilon $ as the smallest value that eliminates multiple predictions for a validation set.

\begin{table*}[ht]
\fontsize{11}{13}\selectfont
\centering
\begin{tabular}{c|c||c|c||c|c||c|c|}
\cline{3-8}
\multicolumn{2}{c|}{}                                                 & \multicolumn{2}{c||}{Estimate $\epsilon$} & \multicolumn{2}{c||}{$\epsilon=0.01$} & \multicolumn{2}{c|}{$\epsilon=0.02$} \\ \hline
\multicolumn{1}{|c||}{Dataset}                    & NC Functions       & $\epsilon$            & Errors           & Errors          & Multiples          & Errors          & Multiples          \\ \hline\hline
\multicolumn{1}{|c||}{\multirow{7}{*}{GTSRB}}     & $k$-NN & 0.021                 & 2.3\%            & 1.1\%           & 3.5\%              & 2\%             & 0.2\%              \\ \cline{2-8} 
\multicolumn{1}{|c||}{}                           & 1-NN               & 0.016                 & 2.3\%            & 1.3\%           & 2.1\%              & 2.4\%           & 0\%                \\ \cline{2-8} 
\multicolumn{1}{|c||}{}                           & Nearest Centroid   & 0.026                 & 2.8\%            & 1.2\%           & 4.8\%              & 2.1\%           & 1.8\%              \\ \cline{2-8} 
\multicolumn{1}{|c||}{}                           & Margin             & 0.034                 & 3.4\%            & 0.9\%           & 7.1\%              & 1.8\%           & 3.4\%              \\ \cline{2-8} 
\multicolumn{1}{|c||}{}                           & Brier Score        & 0.035                 & 3.6\%            & 0.9\%           & 7.1\%              & 2\%             & 3.5\%              \\ \cline{2-8} 
\multicolumn{1}{|c||}{}                           & TS Margin          & 0.034                 & 3.4\%            & 0.7\%           & 8.8\%              & 1.7\%           & 3.7\%              \\ \cline{2-8} 
\multicolumn{1}{|c||}{}                           & TS Brier Score     & 0.035                 & 3.6\%            & 0.7\%           & 8.8\%              & 1.7\%           & 4.1\%              \\ \hline\hline
\multicolumn{1}{|c||}{\multirow{7}{*}{SCITOS-G5}} & $k$-NN & 0.14                  & 16.5\%           & 0.9\%           & 63\%               & 1.3\%           & 51.4\%             \\ \cline{2-8} 
\multicolumn{1}{|c||}{}                           & 1-NN    & 0.092                 & 10.4\%           & 1.8\%           & 43.6\%             & 2.7\%           & 33.5\%             \\ \cline{2-8} 
\multicolumn{1}{|c||}{}                           & Nearest Centroid   & 0.367                 & 35.5\%           & 0.9\%           & 93.9\%             & 1\%             & 93.4\%             \\ \cline{2-8} 
\multicolumn{1}{|c||}{}                           & Margin             & 0.109                 & 13.2\%           & 0.7\%           & 58.6\%             & 1.3\%           & 46.7\%             \\ \cline{2-8} 
\multicolumn{1}{|c||}{}                           & Brier Score        & 0.113                 & 13.4\%           & 0.7\%           & 58.8\%             & 1.3\%           & 47.3\%             \\ \cline{2-8} 
\multicolumn{1}{|c||}{}                           & TS Margin          & 0.109                 & 13.2\%           & 0.9\%           & 52.2\%             & 1.6\%           & 40.8\%             \\ \cline{2-8} 
\multicolumn{1}{|c||}{}                           & TS Brier Score     & 0.113                 & 13.2\%           & 0.9\%           & 52.2\%             & 1.5\%           & 41.8\%             \\ \hline 
\end{tabular}
\caption{Test results for different values of $\epsilon$ }
\label{tab:epsilon_performance}
\end{table*}

Table~\ref{tab:epsilon_performance} shows the results for the datasets and the various nonconformity functions. First using the calibration/validation dataset, we select $\epsilon$ to eliminate sets of multiple predictions and we report the errors in the predictions for the testing dataset. The algorithm did not generate any set with multiple predictions for the testing datasets for any of the nonconformity functions. The nonconformity functions computed using the softmax layer result in a slightly larger error but the results show that the error rates are well-calibrated for all nonconformity functions. The results for the GTSRB dataset exhibit smaller error than the SCITOS-G5 results because the underlying model has much better accuracy.

Table~\ref{tab:epsilon_performance} also reports the results for $\epsilon = 0.01$ and $\epsilon = 0.02$ including the percentage of errors and multiple predictions. For example, in the case of the $k$-NN nonconformity measure and $ \epsilon = 0.01 $ the error is as expected close to $1\%$ but we also have $3.5\%$ predictions with multiple classes. The nonconformity functions computed based on the representations of the penultimate layer result in more efficient predictions. Temperature scaling does not seem to affect the results for nonconformity functions computed using the softmax layer. It should be noted that Hinge nonconformity function did not perform well and the corresponding results are not included in the table. 

For comparison we apply a baseline method using ICP directly on the inputs. In Table~\ref{tab:epsilon_performance_baseline}, we present the results using the same datasets and nonconformity functions. The baseline method requires more sets of multiple predictions to achieve a given confidence level and the significance level $\epsilon$ required to produce single predictions is significantly larger. 

\begin{table*}[ht]
\fontsize{11}{13}\selectfont
\centering
\begin{tabular}{c|c||c|c||c|c||c|c|}
\cline{3-8}
\multicolumn{2}{c|}{}                                                 & \multicolumn{2}{c||}{Estimate $\epsilon$} & \multicolumn{2}{c||}{$\epsilon=0.01$} & \multicolumn{2}{c|}{$\epsilon=0.02$} \\ \hline
\multicolumn{1}{|c||}{Dataset}                    & NC Functions       & $\epsilon$            & Errors           & Errors          & Multiples          & Errors          & Multiples          \\ \hline\hline
\multicolumn{1}{|c||}{\multirow{3}{*}{GTSRB}}     & $k$-NN & 0.198                 & 16.9\%            & 0\%           & 100\%              & 1.8\%             & 76.5\%              \\ \cline{2-8} 
\multicolumn{1}{|c||}{}                           & 1-NN               & ---                 & ---            & ---           & ---              & ---           & ---                \\ \cline{2-8} 
\multicolumn{1}{|c||}{}                           & Nearest Centroid   & 0.825                 & 85.3\%            & 2.5\%           & 100\%              & 3.5\%           & 100\%              \\ \hline\hline
\multicolumn{1}{|c||}{\multirow{3}{*}{SCITOS-G5}} & $k$-NN & 0.198                  & 22.3\%           & 0.7\%           & 72.2\%               & 1.6\%           & 58.4\%             \\ \cline{2-8} 
\multicolumn{1}{|c||}{}                           & 1-NN    & 0.122                 & 12.6\%           & 1.1\%           & 57.9\%             & 3.6\%           & 37.5\%             \\ \cline{2-8} 
\multicolumn{1}{|c||}{}                           & Nearest Centroid   & 0.428                 & 43.5\%           & 0.5\%           & 96.9\%             & 0.7\%             & 95.9\%             \\ \hline 
\end{tabular}
\caption{Test results for different values of $\epsilon$ using the baseline ICP with raw inputs}
\label{tab:epsilon_performance_baseline}
\end{table*}

\subsection{Computational Efficiency}
\label{sec:computation_efficiency}
In order to evaluate if the approach can be used for real-time monitoring of CPS, we measure the execution times and the memory requirements.  
Different nonconformity functions lead to different execution times and memory requirements. We compare the average execution time over the testing datasets for generating a prediction set after the model receives a new test input in Table \ref{tab:time_memory}. The $1$-NN nonconformity function on the input space of the GTSRB dataset has excessive memory requirements. Below we present the computational requirements for each nonconformity function and explain the higher requirements of the $1$-NN function in more detail.

\begin{table*}[ht]
\fontsize{11}{13}\selectfont
\centering
\begin{tabular}{c|c|c||c|c|}
\cline{2-5}
                                         & \multicolumn{2}{c||}{GTSRB} & \multicolumn{2}{c|}{SCITOS-G5} \\ \hline
\multicolumn{1}{|c||}{NC Functions}       & Execution Time  & Memory   & Execution Time    & Memory     \\ \hline
\multicolumn{1}{|c||}{$k$-NN (baseline)} & 81.1ms           & 836.2 MB  & 1.3ms             & 1.7 MB    \\ \hline
\multicolumn{1}{|c||}{$1$-NN (baseline)}    & ---            & ---   & 2.8ms             & 3.4 MB    \\ \hline
\multicolumn{1}{|c||}{Nearest Centroid (baseline)}   & 10.1ms           & 1.1 MB  & 1ms               & 8.8 kB      \\ \hline
\multicolumn{1}{|c||}{$k$-NN} & 6.9ms           & 71.5 MB  & 1.3ms             & 1.13 MB    \\ \hline
\multicolumn{1}{|c||}{$1$-NN}    & 30ms            & 1.4 GB   & 3.1ms             & 4.13 MB    \\ \hline
\multicolumn{1}{|c||}{Nearest Centroid}   & 7.2ms           & 40.7 MB  & 1ms               & 39 kB      \\ \hline
\multicolumn{1}{|c||}{Margin}             & 7ms             & 40.7 MB  & 1.1ms             & 38.3 kB    \\ \hline
\multicolumn{1}{|c||}{Brier Score}        & 6.8ms           & 40.7 MB  & 1.1ms             & 38.3 kB    \\ \hline
\multicolumn{1}{|c||}{TS Margin}          & 7.1ms           & 40.7 MB  & 1ms               & 38.3 kB    \\ \hline
\multicolumn{1}{|c||}{TS Brier Score}     & 6.9ms           & 40.7 MB  & 1.1ms             & 38.3 kB    \\ \hline
\end{tabular}
\caption{Execution Times and Memory Requirements}
\label{tab:time_memory}
\end{table*}

Table~\ref{tab:time_memory} reports the average execution time for each test input and the required memory space using different nonconformity functions. The GTSRB dataset has 19180 training data each represented by an encoding of size 128 while the SCITOS-G5 dataset has 3928 training data each represented by an encoding of size 20. The execution times for the different nonconformity functions are very similar. All the nonconformity functions require storing the trained DNN and the calibration NC scores which are used for computing the test NC scores online. However, it should be noted that the DNN is stored anyway for performing the original task. In the $k$-NN case, the encodings of the training data are stored in a $k-d$ tree~\cite{Bentley:1975:MBS:361002.361007} that is used to compute efficiently the $k$ nearest neighbors. This data structure is used both for the $k$-NN and $1$-NN NC functions. In the $1$-NN case, it is required to find the nearest neighbor in the training data for each possible class which is computationally expensive resulting in larger execution time. The nearest centroid nonconformity function requires storing only the centroids for each class and the additional memory required is minimal.

In conclusion, the evaluation results demonstrate that monitoring based on the conformal prediction framework using embedding representations of the learned models has well-calibrated error rates and can minimize the number of alarms due to predictions with multiple classes. The estimated confidence bounds that will produce sets of single predictions are larger than the baseline ICP application on the inputs. Further, the approach allows selecting the significance level to trade-off errors and alarms. Finally, the use of the embedding space reduces the memory requirements and the execution time when the nonconformity function needs to have access to the whole dataset which justifies the use of ICP in the learned embedding space.
\section{Concluding Remarks}
\label{sec:conclusion}
Cyber-physical systems (CPS) incorporate machine learning components such as DNNs for performing various tasks such as perception of the environment.
Although DNNs offer advanced capabilities, they must be complemented by engineering methods and 
practices that allow effective integration in CPS. 
The paper considers the problem of complementing the prediction of DNNs
 with a computation of confidence. For classification tasks, 
in addition to the class prediction, we compute set and confidence predictors using 
 the conformal prediction framework and we present computationally efficient algorithms based on representations learned by the underlying model that can be used for real-time monitoring. 
 We perform an empirical evaluation of the approach using a traffic sign recognition benchmark and a robot navigation dataset.
The evaluation results demonstrate that monitoring based on the conformal prediction framework using embedding representations of the learned models has well-calibrated error rates and can minimize the number of alarms due to predictions with multiple classes. Further, the approach allows selecting the significance level to trade-off errors and alarms. Finally, the approach is computationally efficient and can be used for real-time monitoring of CPS.
\section*{Acknowledgements}
\label{sec:acknowledgements}
The material presented in this paper is based upon work supported by the National Science Foundation (NSF) under grant numbers CNS 1739328,  the Defense Advanced Research Projects Agency (DARPA) through contract number FA8750-18-C-0089, and the Air Force Office of Scientific Research (AFOSR) DDDAS through contract number FA9550-18-1-0126. The views and conclusions contained herein are those of the authors and should not be interpreted as necessarily representing the official policies or endorsements, either expressed or implied, of AFOSR, DARPA, or NSF.

\end{document}